\documentclass[12pt,draftclsnofoot,onecolumn]{IEEEtran}
\usepackage{amsmath,amsfonts}
\usepackage[linesnumbered,ruled,vlined]{algorithm2e}

\usepackage{array}
\usepackage[caption=false,font=normalsize,labelfont=sf,textfont=sf]{subfig}
\usepackage{textcomp}
\usepackage{stfloats}

\usepackage{url}
\usepackage{verbatim}
\usepackage{graphicx}
\usepackage{amsmath}
\usepackage{bbm}
\usepackage{helvet}
\usepackage{courier}
\usepackage{amssymb}
\usepackage{cite}
\usepackage{algpseudocode}

\begin{document}

\title{Combat Urban Congestion via Collaboration: Heterogeneous GNN-based MARL for Coordinated Platooning and Traffic Signal Control }
\author{Xianyue Peng,
Shenyang Chen,
Hang Gao,
Hao Wang,
H. Michael Zhang}

% 在 maketitle 之前插入声明（单栏草稿版用这个）
\IEEEaftertitletext{%
\par\noindent\begin{center}
\footnotesize
© 2025 IEEE. Personal use of this material is permitted. Permission from IEEE must be obtained for all other uses.\\
The final version of record is available at: \url{https://ieeexplore.ieee.org/document/10977660}\\
DOI: 10.1109/TITS.2025.10977660
\end{center}\par\vspace{1ex}
}


\maketitle
\begin{abstract}
Over the years, reinforcement learning has emerged as a popular approach to develop signal control and vehicle platooning strategies either independently or in a hierarchical way. However, jointly controlling both in real-time to alleviate traffic congestion presents new challenges, such as the inherent physical and behavioral heterogeneity between signal control and platooning, as well as coordination between them. This paper proposes an innovative solution to tackle these challenges based on heterogeneous graph multi-agent reinforcement learning and traffic theories. Our approach involves: 1) designing platoon and signal control as distinct reinforcement learning agents with their own set of observations, actions, and reward functions to optimize traffic flow; 2) designing coordination by incorporating graph neural networks within multi-agent reinforcement learning to facilitate seamless information exchange among agents on a regional scale; 3) applying alternating optimization for training, allowing agents to update their own policies and adapt to other agents' policies. We evaluate our approach through SUMO simulations, which show convergent results in terms of both travel time and fuel consumption, and superior performance compared to other adaptive signal control methods.
\end{abstract}

\section{Introduction}

\IEEEPARstart{U}{rban} population growth has intensified traffic congestion, leading to longer commutes, higher fuel consumption, and increased greenhouse gas emissions. A 2022 INRIX study \cite{INRIX} found that an average US driver lost 99 hours and $\$1,325$ in $2022$ due to congestion, versus $\$1,010$ in 2021 in the United States (US). In 2020, \cite{Fast_Facts} reported that $27\%$ of greenhouse gas emissions stemmed from transportation, with most coming from light and medium-duty vehicles. These \IEEEpubidadjcol rising challenges highlight the urgent need for innovative solutions to optimize transportation systems. Traffic signal control and platooning control are widely acknowledged as effective strategies for mitigating traffic congestion and optimizing traffic flow within urban road networks. Platooning control involves managing the timing of platooned vehicles arriving at intersections, while signal control determines signal timings based on the spatiotemporal distribution of approaching traffic. However, conventional platooning and signal control heavily rely on simplified models with various assumptions about driver behaviors and are often complex to optimize when applied to large networks. Deep reinforcement learning (DRL) is shown to handle complex tasks in real-world decisions with ease and is model-free. It has gained attention for applications in in traffic signal control, autonomous driving, and platoon control \cite{chen2020toward,xu2021hierarchically,yang2021ihg,yen2022deep}.

Despite the progress in RL-based research on intelligent transportation systems, there's limited focus on integrating platoon control with signal control in a Multi-agent DRL framework. It is imperative to recognize that platoon control and signal control exhibit a notable blend of physical and behavioral heterogeneity, which increases the complexity as a heterogeneous multi-agent system problem. Furthermore, the coordination between these two controls brings a new control dimension. Platooning and signal control agents formulate decisions based on individual policies through local observations, giving rising to non-cooperative actions with suboptimal outcomes on a global scale.

To tackle these challenges, we propose a cooperative RL-based platooning and signal control method, which is called JointSP. The work's main contributions can be summarized as follows:
\begin{itemize}
    \item Multi-Agent Framework: The approach employs a multi-agent framework to coordinate the activities of platoons and signal controls. Platoon Agents (PAs) manage platoon maneuvers, while Signal Control Agents (SAs) optimize signal timings at intersections. Together, platoons can be guided to traverse signalized intersections with minimal or no stops, leading to a notable enhancement in overall traffic throughput and efficiency.
    \item SA and PA Design: 
    Due to the physical and behavioral heterogeneity, both SAs and PAs have been carefully structured, encompassing their observations, actions, and reward components. Recognizing the diversity across SAs at different intersections, our approach customizes each SA for its geographic characteristic and enables a \textit{Decentralized Training Decentralized Execution} (DTDE) learning process. In contrast, due to the shared policies and objectives among PAs, they undergo a \textit{Centralized Training Decentralized Execution} (CTDE) learning procedure.
    \item Heterogeneous Graph Neural Network Multi-Agent Reinforcement Learning (HGMARL): To tackle the complexities of multi-agent systems, JointSP integrates graph neural networks (GNNs) into multi-agent reinforcement learning. The interplay between PAs and SAs shapes the GNN structure, facilitating efficient information exchange and fostering collaborative decision-making among these agents.
    \end{itemize}
In summary, this work innovatively combines MADRL and heterogeneous GNN to enhance the coordination and decision-making capabilities of the platoon and signal control agents, ultimately improving traffic management in complex traffic conditions.

\section{Related Work}
\subsection{RL based traffic signal control}
In recent years, DRL has found extensive applications in traffic signal control. \cite{liang2019deep} employed DRL for traffic light cycle control. 
Addressing high-dimensional state and action spaces, \cite{lee2019reinforcement} introduced a reinforcement learning algorithm to manage the entire traffic state and signal network simultaneously. Viewing distributed signal control networks as multi-agent systems, researchers have utilized the MADRL algorithm. \cite{abdoos2021hierarchical} designed a hierarchical multi-agent system for extensive traffic signal control. First-level agents employ reinforcement learning to identify optimal policies, while second-level agents use LSTM neural networks to estimate traffic states. Incorporating an edge computing framework, \cite{zhang2021cooperative} devised a cooperative multi-agent actor-critic DRL technique. Their method integrates local agent contribution weights for global optimization in traffic control. \cite{yang2021ihg} proposed a decentralized MADRL approach with an inductive heterogeneous graph neural network to address multi-intersection signal control.
\subsection{RL based platooning control}
The utilization of reinforcement learning (RL) in platooning control has emerged as a promising avenue for optimizing traffic performance and enhancing road safety. By harnessing RL's capacity for non-linear approximations and unsupervised learning, several approaches have been introduced to design platoon control strategies that improve energy efficiency and road capacity. \cite{desjardins2011cooperative} pioneered the application of RL to Cooperative Adaptive Cruise Control (CACC) with a model-free algorithm managing basic cruise control actions within a platoon. Subsequent studies have advanced this concept, with \cite{buechel2018deep} incorporating predictive vehicle trajectories for longitudinal control, \cite{chu2019model} developed a model-based deep RL algorithm for heterogeneous platoons, and \cite{lu2021sharing} using shared models to address exploration challenges within platoon control. \cite{chen2020platooning} devised a DRL algorithm for optimizing platoon maneuvers and entry points, while \cite{li2017training} introduced RL-based adaptive cruise control to enhance fuel efficiency and safety. To counteract issues like traffic oscillations, \cite{zhou2019development} proposed a DDPG-based approach for controling acceleration, further expanded upon by \cite{lei2022deep} with a Finite-Horizon DDPG framework. \cite{shi2023deep} introduced a general platooning framework employing DPPO-based algorithms to manage mixed traffic patterns and stabilize traffic oscillations. Collectively, these studies showcase the versatility and potential of RL in revolutionizing platooning control strategies.
\subsection{Coordinated RL based platooning and signal control}
Research on coordinated platooning and signal controls remains relatively scarce. \cite{berbar2022reinforcement} introduced a double agent reinforcement learning method for an isolated signalized intersection, training the Velocity Agent to manage both platoon and individual Connected and Autonomous Vehicle (CAV) speeds, followed by training the Signal Agent to improve traffic flow efficiency through signal sequencing and phasing. However, this study lacks consideration of the collaborative relationship between vehicle agents and signal agents, training them separately, thus yielding limited synergy. Moreover, the study models all vehicles as a single Velocity Agent and confines the Signal Agent to a single isolated intersection, thereby constraining the model's scalability to handle broader problem scenarios.

In addition to joint platooning control and signal control, other vehicular cooperative scenarios have also utilized reinforcement learning algorithms as references for our work. For instance, in \cite{sun2023hierarchical}, a decentralized bi-directional hierarchical reinforcement learning framework was developed to jointly control traffic signal plans and rerouting of autonomous vehicles in mixed traffic scenarios.

Research on cooperative multi-agent reinforcement learning (MARL) methods remains a vibrant area of rexploration. \cite{lowe2017multi} proposed the integration of deep reinforcement learning techniques into multi-agent domains, effectively overcoming traditional algorithmic limitations by adopting adaptations like actor-critic methods and ensemble learning. These adaptations enhance coordination strategies among agent populations. Subsequently, \cite{wang2020roma} introduced the Role-oriented MARL framework (ROMA), seamlessly merging role-based design with multi-agent reinforcement learning. This innovative framework dynamically allows roles to emerge, facilitating agents with shared roles to collaboratively specialize in specific tasks, thereby offering a flexible approach for complex multi-agent systems. Addressing parameter sharing, \cite{christianos2021scaling} delved into the nuanced issue within multi-agent deep reinforcement learning, proposing a method that intelligently identifies agents that genuinely benefit from shared parameters based on their individual capabilities and objectives. \cite{bettini2023heterogeneous}, the most recent contribution, confronted the demand for policy heterogeneity in traditional MARL frameworks. This study introduced the HetGPPO method, utilizing GNNs for inter-agent communication to enable diverse behavior learning and effective collaboration in partially observable environments. These collective contributions advance our understanding of cooperative MARL, offering an array of innovative methodologies that address coordination, agent parameter sharing, and heterogeneity concerns. This paper's algorithm combines these achievements, integrating dynamic agent emergence, parameter sharing, and agent heterogeneity while applying them to coordinated platooning and signal controls, a prototypical cooperative control application.

\section{Preliminaries}
This paper aims to investigate the joint control of platooning and traffic signals within a city traffic network. Before introducing the RL model, we first introduce the multi-agent extension of a Markov Decision Process (MDP), then characterize the involved mechanisms in terms of the two control units.

\subsection{Markov Games}
Define a tuple 
\begin{equation}
\langle \mathcal{V}, \mathcal{S}, \mathcal{O}, \mathcal{A}, \{\mathcal{R}_i\}_{i \in \mathcal{V}, }, \mathcal{T},
\gamma \rangle
\end{equation}
where $\mathcal{V} = \{ 1,...,n \}$ denotes the set of agents, $\mathcal{S}$ is the state space, $\mathcal{O} = \mathcal{O}_1 \times ... \times \mathcal{O}_n$ and $\mathcal{A} = \mathcal{A}_1 \times ... \times \mathcal{A}_n$ are the observations and actions spaces, with $\mathcal{O}_i \in \mathcal{S}$, $\forall i \in \mathcal{V}$ and  $\{\mathcal{R}_i\}_{i \in \mathcal{V}}$ are agent observations and reward functions. In a fully observable MDP, agents observe the true state of the environment, such that $\mathcal{O}_i = \mathcal{S}_i$. Reward is a function of state and actions $\mathcal{R}_i$: $\mathcal{R} \times \mathcal{A} \times \mathcal{S} \rightarrow \mathbb{R}$. $\mathcal{T}$ is the stochastic state transition model, defined as $\mathcal{T}$: $\mathcal{S} \times \mathcal{A} \times \mathcal{S} \rightarrow [ 0,1 ]$. Lastly, $\gamma$ is the discount factor.

In a communication graph $\mathcal{G} = (\mathcal{V}, \mathcal{E})$, node $i \in \mathcal{V}$ represents agents and edge $e_{i,j,t} \in \mathcal{E}$ represents communication links. The communication connection of each agent is defined as $\mathcal{N}_{i,t} = \{ v_j \in \mathcal{V} |e_{j,i,t} \in \mathcal{E} \}$. At each timestep $t$, each agent $i$ gets an observation $o_{i,t}$. We further define $s_{i,t} = \{ o_{i,t}\}\cup\{ o_{j,t} | j \in \mathcal{N}_{i,t}\}$ as observations from itself and its connected agents $\mathcal{N}_i$. A (stochastic) policy $\pi_i$ uses this information to compute an action $a_{i,t} \backsim \pi_i(\cdot |s_{i,t})$, where $a_t = [a_{1,t},...,a_{n,t}] \in \mathcal{A}$. Then actions are implemented in the transition model to obtain the next state $s_{t+1} \backsim \mathcal{T}(\cdot | s_t, a_t)$. A reward $r_{i,t} = \mathcal{R}_i(s_t, a_t, s_{t+1})$ is received for agent $i$. In a finite-horizon ($T$) POMDP, the goal of each agent is to maximize the sum of discounted rewards $\mathcal{R}_{i,t} = \sum_{k=0}^T \gamma^k r_{i,{t+k}}$, which is called the return. An agent aims to maximize the expected discounted return by finding a good policy $\pi_i$.

\subsection{Signal Control System}
In the road network, numerous signalized intersections are distributed, and a signal control system is designed for the network to regulate the signal timings of each intersection. This paper adopted phase selection as the signal control strategy, entailing the selection of a specific phase for each intersection at every control step. A traffic signal phase $p$ is a set of allowable traffic movements. In other words, one phase is selected during each control step, and the corresponding movements for vehicles are permitted to proceed. An approach lane $l$ is defined as the entry lane to an intersection, where each approach lane may accommodate the same or different traffic flow movements, depending on the specific layout of the intersection. 

\begin{figure}[h]
	\centering
		\includegraphics[scale=.35]{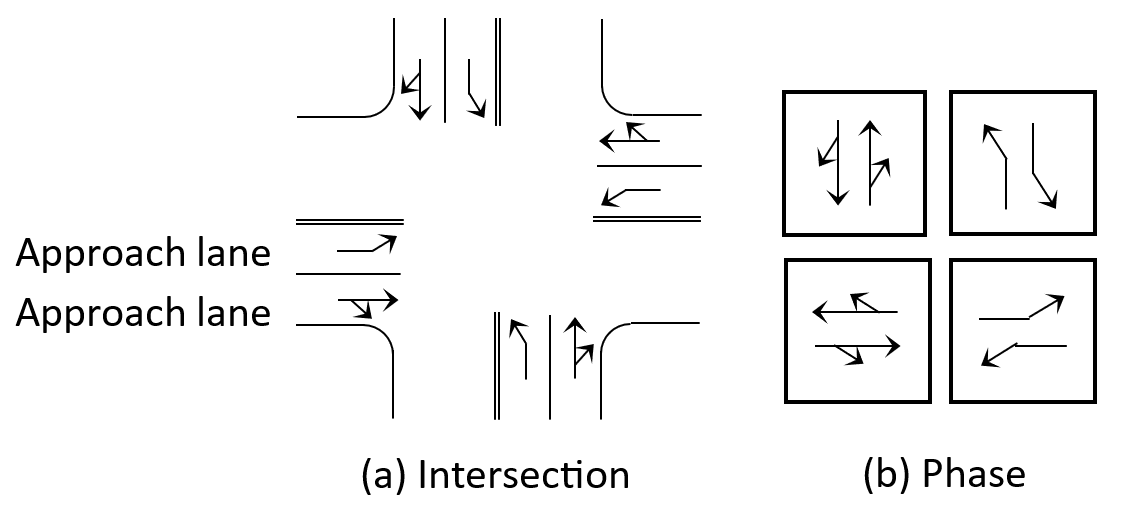}
	\caption{Illustration of the components of an intersection}
	\label{fig: intersection}
\end{figure}

\subsection{Platooning Maneuvers}
\label{subsection: platoon maneuvers}
In a CAV environment, vehicles can leverage wireless communications to organize the formation and maneuvers of platoons. These platoons, dynamically managed, typically consist of a leading vehicle along with several followers. The leading vehicle sets the cruising speed, thereby dictating the pace for the entire platoon.  In response, the following vehicles adjust their driving behaviors to stay aligned with the leading vehicle’s pace and trajectory.

Within the platoon, the leading vehicle travels as fast as possible within the speed limit while ensuring safety; the following vehicles closely follow the leading vehicle with ideal following behavior. To describe the above behaviors of vehicles and to meet the basic requirements of platoon driving strategy, it is assumed that the leading vehicle follows the Krauss model (Krauß, 1998), which emphasizes maximizing speed while ensuring safety. In contrast, the following vehicles employ the Enhanced Intelligent Driver Model (EIDM) developed by (Salles et al., 2020), which refines the Intelligent Driver Model (IDM). The IDM model assumes an instantaneous driver response without accounting for reaction time delays, necessitating that following vehicles dynamically adjust their speed and acceleration based on the leading vehicle's behavior to maintain a safe distance. Building on the IDM model, the EIDM model more precisely controls the micro-acceleration of vehicles to prevent abrupt changes in speed.

In the system, there are platoon vehicles, including a leader and followers; and non-platoon vehicles, namely solo vehicles. Fig. 2 illustrates the dynamic processes of merging and splitting within platoons. Vehicles merge into a unified platoon when the time headway or distance between them falls below a specified threshold. This merging process involves three scenarios: "Formation," where two or more solo vehicles merge into a platoon; "Extension," where the platoon incorporates vehicles traveling behind it; and "Merge," where two platoons combine into a larger platoon. On the other hand, platoon splitting occurs when vehicles need to separate due to differing subsequent travel paths, typically involving the leader leaving and followers leaving. When the leader leaves, the followers become solo vehicles, awaiting further steps to form a new platoon; when a follower leaves, the vehicles ahead maintain platoon formation, while those behind revert to solo status, also waiting for further steps to reorganize into a new platoon.

Typically, vehicles select the highest feasible speed based on current environmental conditions. However, at signalized intersections, allowing vehicles that cannot pass through the intersections to decelerate in advance can save fuel and provide smoother speed transitions. The platoon management system offers suggested speeds for platoons for this purpose. Specifically, as illustrated in Fig. 3, if the platoon can pass the intersection at the current speed, it maintains a steady speed through the intersection. If the platoon cannot pass the intersection at the current speed, it decelerates in advance and then either passes through the intersection or stops to wait for the green light.

\begin{figure}
	\centering
		\includegraphics[scale=.7]{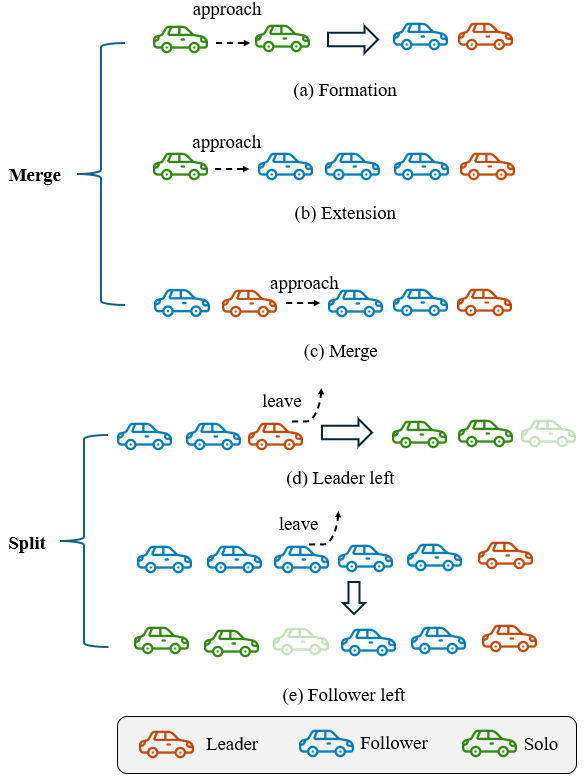}
	\caption{Platoon merging and splitting}
	\label{fig: Platoon Merging and Splitting}
\end{figure}

\begin{figure}
	\centering
		\includegraphics[scale=.25]{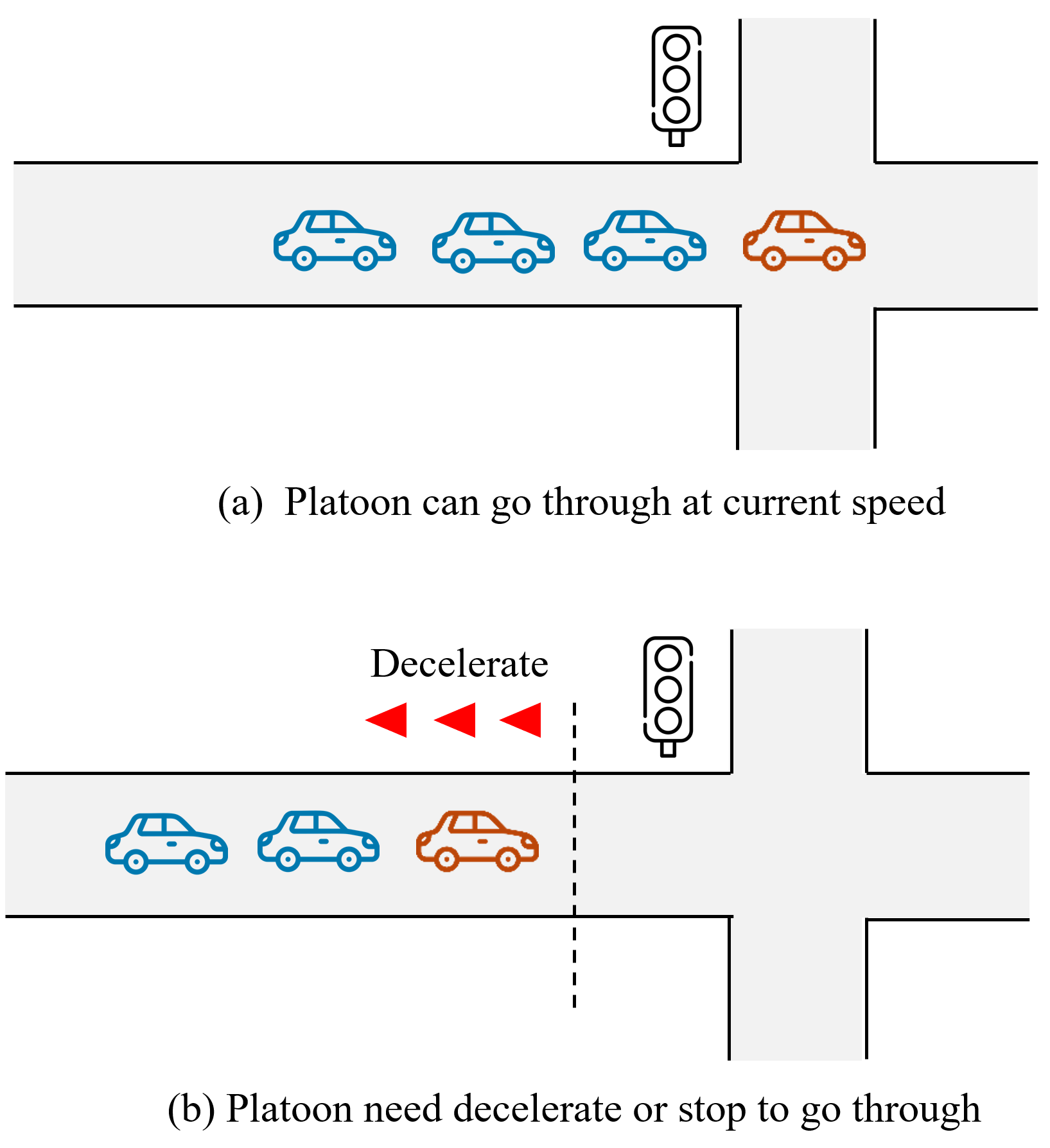}
	\caption{Platoon maneuvers at intersections}
	\label{fig: platoon maneuvers at a signalized intersection}
\end{figure}

\section{Methodology}
Define signal agent and platooning agent as SA and PA, respectively.  In this section, we first characterize the homogeneity and heterogeneity among PAs and SAs. Then we define their specified observations, actions, and reward functions. Finally, we present the JointSP model and show its learning process.

\subsection{Homogeneity and Heterogeneity}
According to \cite{bettini2023heterogeneous}, the definition of heterogeneous systems differs in physics, behaviors, and objectives. But are identical to homogeneous systems. In our multi-agent framework, we clarify the homogeneity and heterogeneity as follows:

\begin{itemize}
    \item  PA and SA display both physical and behavioral heterogeneity, evident through their unique observations and distinct policies, which have been illustrated in the preliminaries. They also do not necessarily share the same objective to accomplish coordination. For example, PAs may focus on maintaining a steady speed to travel through the network \cite{yen2022deep}, whereas SAs focus on reducing waiting time and traffic travel delay around intersections.
    \item Due to varying intersection layouts, SAs are physically and behaviorally different. This diversity results in distinct observations (collected from approach lanes) and selections of actions (representing potential signal phases). But all SAs can share a common objective, such as minimizing waiting time.
    \item PAs are physically identical, sharing the same behavioral models and objectives. They are intrinsically homogeneous.
\end{itemize}

According to the above characteristics, homogeneous PAs share a common policy and corresponding neural network for decision-making. Conversely, each SA possesses an independent policy and corresponding neural network.

\subsection{Signal Control Agent}
Signal control agent (SA) optimizes the signal timings of an intersection, thereby reducing vehicle delay. The observation, action and reward definitions for SA are defined as follows.

\subsubsection{Observation Space $\mathcal{O}_{\rm SA}$}
We consider the observation of SA as a set of indicators for each approach lane at the intersection. Specifically, the observation of SA is defined as
\begin{equation}
o_{i,t} = q_{i,t} \quad \forall i \in \mathcal{V}_{\rm SA}
\end{equation}
where $\mathcal{V}_{\rm SA}$ is the set of SAs, $o_{i,t}$ is the observation for agent $i$ at control step $t$. $q_{i,t} = \begin{bmatrix} q_{i,l,t} \end{bmatrix}_{l\in L_i}$, where $q_{i,l,t}$ represents the number of queued vehicles in approach lane $l$ for agent $i$ at control step $t$. $L_i$ denotes the set of the approach lanes of intersection $i$.

\subsubsection{Action Space $\mathcal{A}_{\rm SA}$}
The definition of heterogeneous SAs allows for DTDE. Each SA $i$ selects an action $a_{i,t}$ from its own feasible action set $\mathcal{A}_{{\rm SA}_i} = P_i$, which are predetermined by intersection layouts. Together, we have
\begin{equation}
a_{i,t} \in \mathcal{A}_{{\rm SA}_i}\quad \forall i \in \mathcal{V}_{\rm SA}
\end{equation}

\subsubsection{Reward $\mathcal{R}_{\rm SA}$} 
We define the reward of SA as the total waiting time during the control step and the waiting time of the first vehicle since it stops. The second term deal with excessive delays and fairness with respect to delays.
\begin{equation}
r_{i,t} = -\sum_{l \in L_i} \left( w_{i,l,t} + \alpha f_{i,l,t} \right) \quad \forall i \in \mathcal{V}_{\rm SA}
\end{equation}
where $ w_{i,l,t}$ represents the waiting time in approach lane $l$ for agent $i$ during control step $t$. $f_{i,l,t}$ represents the waiting time of the first vehicle in the queue since it stops, and $\alpha$ is a coefficient.

\subsection{Platoon Agent}
One platoon agent controls a platoon in the system. Since all platoon agents share the same policies, the system can handle hundreds of agents simultaneously without a substantial increase in computational complexity.

Unlike SAs that exist throughout the entire process, PAs form and disband at different times. In the initial state, all vehicles in the system are solo vehicles and some of these vehicles may form platoons during the simulation. 

As described in Section III.C, vehicles form a platoon (Fig. 2(a)) when the headway or distance between them falls below a specified threshold. A platoon agent is then allocated for decision-making for the platoon. During the decision-making process, the platoon may extend by incorporating following solo vehicles (Fig. 2(b)) or following platoons (Fig. 2(c)), or split when a follower leaves (Fig. 2(e)).

The platoon agent reaches its terminal state when the leader of the platoon changes lanes, enters another link, exits the network (Fig. 2(d)), or merges with another platoon ahead (Fig. 2(c)).

At each control step, a platoon agent provides driving speed recommendations to a platoon to ensure they pass through a set of intersections efficiently while optimizing fuel consumption. The observation, action, and reward definitions are defined below.

\subsubsection{Observation Space $\mathcal{O}_{\rm PA}$}
The observation of each PA at time step $t$ includes the characteristics of the platoon including platoon size, speed and position, as well as the signal timings. 
\begin{equation}
o_{i,t} = (d_{i,t},n_{i,t},v_{i,t}, p_{r_i,t},p_{i,t}) \quad \forall i \in \mathcal{V}_{\rm PA}  
\end{equation}
where $d_{i,t}$ denotes the distance from the leading vehicle of $i^{th}$ PA to the approaching signalized intersection $r_i$; $n_{i,t}$ represents the number of vehicles in $i^{th}$ PA; $v_{i,t}$ represents the speed of the platoon; $p_{r_i,t}$ represents the current phase index at signalized intersection $r_i$; $p_{i,t}$ represents the phase index of intersection \( r_i \) when the platoon is allowed to pass.

\subsubsection{Action Space $\mathcal{A}_{\rm PA}$}
The definition of homogeneous PAs allows for CTDE, which suggests that although different PAs are trained with parameter sharing, they apply distinct actions based on their partially observable state $o_{i,t}$. 

To simulate real traffic, the system employs car-following models to guide vehicle behavior, typically accounting for factors such as safe time headway. Our PAs modulate the platoon's speed by limiting the maximum allowable speed instead of directly controlling vehicle acceleration or deceleration. Consequently, we define their action space as the maximum speed set, \(\mathcal{A}_{\rm PA}\).
\begin{equation}
a_{i,t} \in \mathcal{A}_{\rm PA} \quad \forall i \in \mathcal{V}_{\rm PA}  
\end{equation}

\subsubsection{Reward $\mathcal{R}_{\rm PA}$} 
Since the goal of the MARL algorithm is to maximize cumulative rewards, we must prevent agents from artificially extending the platoon's existence to increase their cumulative rewards. To address this, we assign a reward only when the PA reaches its terminal state, using this reward to evaluate the platoon's performance during its existence. The multi-objective reward, incorporating distance traveled per unit time and fuel consumption per unit distance, is defined as
\begin{equation}
r_{i,t} = 
\begin{cases} 
0 & \text{if } b_{i,t} = 0 \\
\frac{\sum_{\tau=t_i}^{t} x_{i,\tau}}{t - t_i+1} - \beta \frac{\sum_{\tau=t_i}^{t} g_{i,\tau}}{\sum_{\tau=t_i}^{t} x_{i,\tau}} & \text{if } b_{i,t} = 1 
\end{cases}
\quad \forall i \in \mathcal{V}_{\rm PA}
\end{equation}
where $b_{i,t}$ is the termination indicator of the $i^{th}$ PA, indicating whether it has reached its terminal state. $t_i$ is the initial control step of the $i^{th}$ PA. $g_{i,\tau}$, $x_{i,\tau}$ represent the fuel consumption and distance traveled of the $i^{th}$ PA at control step $\tau$, respectively. $\beta$ is a coefficient.

\subsection{JointSP Model}
We propose the Heterogeneous Graph neural network Multi-Agent Reinforcement Learning (HGMARL) model, designed specifically to facilitate the coordination between SAs and PAs. The basic idea of JointSP is to implement the GNN communication layer into the MARL training frameworks. This design promotes efficient information sharing among SAs and PAs, addressing challenges posed by partial observability and real-time coordination. However, it's imperative to realize that: 1) The observations of SAs and PAs have different dimensions. 2) In the transportation systems, platoon trajectories are dynamic, which indicates PAs may interact with different SAs based on their positions, increasing the difficulty of coordination and leading to non-stationarity. Such complexities call for a careful design of our HGMARL model.

\begin{figure}
	\centering
		\includegraphics[scale=.38]{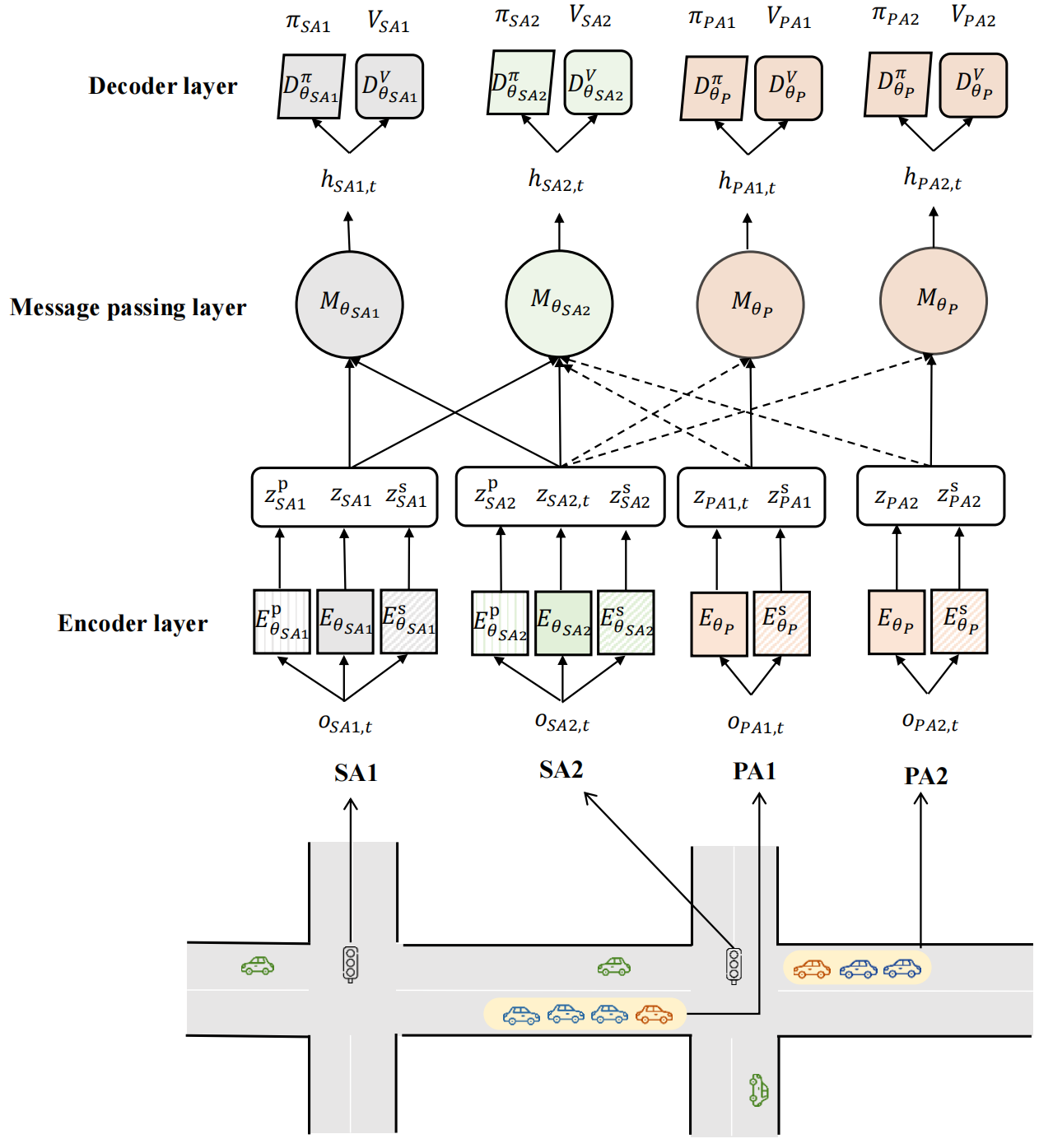}
	\caption{HGMARL framework}
	\label{fig: GNN}
\end{figure}

We depict the model in Fig. \ref{fig: GNN}. Suppose there are 2 SA and 2 PAs. During each control step $t$, each agent $i$ obtains its observation $o_{i,t}$. To tackle the first issue,  we process them through an encoder to generate the embedded observations. Mathematically, it can be defined as
\begin{equation}
z_{i,t} = E_{\theta_i}(o_{i,t})\quad \forall i \in  \mathcal{V}_{\rm SA}\cup \mathcal{V}_{\rm PA}  
\end{equation}
\begin{equation}
z_{i,t}^\mathrm{s} = E^\mathrm{s}_{\theta_i}(o_{i,t})\quad \forall i \in \mathcal{V}_{\rm SA}\cup \mathcal{V}_{\rm PA}  
\end{equation}
\begin{equation}
z_{i,t}^\mathrm{p}= E_{\theta_i}^\mathrm{p}(o_{i,t})\quad \forall i \in \mathcal{V}_{\rm SA}  
\end{equation}
where $E_{\theta_i}(\cdot),E^\mathrm{s}_{\theta_i}(\cdot),E^\mathrm{p}_{\theta_i}(\cdot)$ are encoder layers of agents for itself, its relevant SA, its relevant PA, which are multi-layer perceptron (MLP), parameterized by agent parameter $\theta_i$. $z_{i,t},z_{i,t}^\mathrm{s},z_{i,t}^\mathrm{p}$ are embedded observations, that is, the output of those encoder layers.

The GNN is structured with edges that establish vital and dynamic connections between SAs and PAs. The edge feature $e_{ij}$ builds the connection between agent $i$ and $j$ to tackle the coordination issue. The coordination is twofold: 1) PA builds a link with the approaching SA; 2) SA builds links with neighboring PAs and SAs. Since the number of relevant PAs is dynamic for each SA, we consistently select $n$ closest PAs as input for the GNN layer of SA. Also, unlike PAs with dynamic movements, SAs maintain fixed positions. This means when a SA builds connections with neighboring SAs, the edge features among themselves are unchanged, solely dependent on the traffic network layout. These configurations ensure the robustness of the GNN's architecture in the dynamic traffic system.
Take ${\rm PA}_1$ and ${\rm SA}_1$ as examples shown in Fig. \ref{fig: GNN}. For ${\rm SA}_2$, the neighboring ${\rm SA}$ is ${\rm SA}_1$; the relevant PAs are ${\rm PA}_1$, and ${\rm PA}_2$.  We thus take 
\begin{equation}
e_{{\rm SA}_1,{\rm SA}_2,t} = e_{{\rm PA}_1,{\rm SA}_2,t} = e_{{\rm PA}_2,{\rm SA}_2,t} = 1
\end{equation}
Similarly for ${\rm PA}_1$ , the relevant SA is ${\rm SA}_2$, which means
\begin{equation}
e_{{\rm SA}_2,{\rm PA}_1,t} = 1
\end{equation}
As a result, we can formally define the shared observations for agent $i$ as 
\begin{equation}
s_{i,t} = \{o_{i,t}\}\cup\{o_{j,t} |e_{j,i,t} = 1\}
\end{equation}
The message-passing GNN layer can be formulated as
\begin{equation}
h_{i,t} = M_{\theta_i}\left([z_{i,t}; z_{j,t}^{\rm s} \mid e_{j,i,t} = 1]\right) \quad \forall i \in  \mathcal{V}_{\rm SA} 
\end{equation}
\begin{equation}
h_{i,t} = M_{\theta_i}\left([z_{i,t}; z_{j,t}^{\rm p} \mid e_{j,i.t} = 1]\right) \quad  \forall i \in  \mathcal{V}_{\rm PA} 
\end{equation}
where all associated embedded observations will be concatenated, then pass through the GNN communication layer $M_{\theta_i}(\cdot)$ to produce output $h_{i,t}$.

In the final stage, the model employs two decoder layers. Taking $h_{i,t}$ as input, they derive the policy and value networks, denoted as $\pi_i$ and $V_i$, respectively. They leverage two MLPs to output actions for PAs and SAs.

Finally, the decoder is constructed to obtain the corresponding policy and state value by inputting $h_{i,t}$ into the decoder.
\begin{equation}
\pi_{\theta_i}(a_{i,t} |s_{i,t}) = D^{\pi}_{\theta_i}(h_{i,t}) \quad  \forall i \in  \mathcal{V}_{\rm SA}  \cup  \mathcal{V}_{\rm PA}  
\end{equation}
\begin{equation}
V_{\theta_i}(s_{i,t}) = D^V_{\theta_i}(h_{i,t}) \quad  \forall i \in \mathcal{V}_{\rm SA}  \cup \mathcal{V}_{\rm PA} 
\end{equation}
where $D^{\pi}_{\theta_i}(\cdot),D^V_{\theta_i}(\cdot)$ is the policy decoder layer and value decoder layer of agent $i$, which are MLPs; $\pi_{\theta_i}(a_{i,t} |s_{i,t})$ is the policy function of agent $i$, indicating the probability of choosing action $a_{i,t}$ given state $s_{i,t}$ under parameter $\theta_i$; $V_{\theta_i}(s_{i,t})$ is the value function of agent $i$, representing the expected return starting from state $s_{i,t}$ under parameter $\theta_i$.

Overall, the multi-agent graph neural network includes encoder layers \(E_{\theta_i}(\cdot)\), \(E_{\theta_i}^{\rm s}(\cdot)\), \(E_{\theta_i}^{\rm p}(\cdot)\), message passing layer \(M_{\theta_i}(\cdot)\), and decoder layers \(D_{\theta_i}^{\pi}(\cdot)\) and \(D_{\theta_i}^{V}(\cdot)\). These neural networks are determined by the parameters \(\theta_i\). Due to the homogeneity among PAs and heterogeneity among SAs, the PAs share the same neural network parameters while each SA has independent neural network parameters, which means:

\begin{align}
    \theta_1 = \theta_2 = ... = \theta_i = \theta_{\rm PA}\quad \forall i \in \mathcal{V}_{\rm PA}\\
        \theta_1 \neq \theta_2  \neq \theta_3 \neq ...  \neq \theta_{i} \quad \forall i \in \mathcal{V}_{\rm SA}
\end{align}
\subsection{Training}
We choose PPO \cite{schulman2017proximal} to carry out the training, motivated by the following considerations. 1) Stability: The diversity between SA and PA makes the environment non-stationary and dynamic. PPO addresses these by incorporating a clipped objective, ensuring controlled magnitudes for policy updates. 2) Flexibility: The JointSP model simultaneously includes CTDE and DTDE paradigms, while PPO can be adjusted to both. Our JointSP model has been customized using PyTorch, and the RLlib framework has been utilized for training. 

In the system, there are two types of agents, SAs and PAs. They obtain observations \( o_{i,t} \) and edge features $e_{j,i,t}$ in the GNN from the environment to derive their state \( s_{i,t} \). Each agent selects an action \( a_{i,t} \) based on their respective policies \( \pi_{\theta_{{\rm SA}_i}}(a_{i,t} | s_{i,t}) \) and \( \pi_{\theta_{\rm PA}}(a_{i,t} | s_{i,t}) \). The joint actions \( a_{t}= (a_{i,t})_{i \in \mathcal{V}_{\rm SA} \cup \mathcal{V}_{\rm PA}}\) of multiple agents are executed in the environment, resulting in rewards and the next state. This process is executed for a batch size number of steps, and the transitions are stored in a batch buffer.

Given the batch buffer, the training of SAs and PAs uses alternating optimization. This approach ensures that each agent adapts to the evolving strategies of other agents while maintaining stability and convergence\cite{feng2024hierarchical}. It is essential because the state and reward of SAs and PAs are influenced by the policies of the other type of agent. Simultaneous training could confuse agents due to the non-stationarity problem, where the environment's dynamics change as each agent updates its policy. By fixing the policy of one type and updating the other, agents can indirectly learn the strategy of the other type, leading to better predictions. For example, fixing SA policies and updating PA policies with SA states through the GNN enables PAs to predict signal timings more accurately, resulting in better speed choices. 

\begin{algorithm}
\caption{Environment Iteration}\label{algorithm1}
\fontsize{9pt}{10pt}\selectfont
    \For{t in range(batch\_size)}{
        \ForEach{agent $i \in \mathcal{V}_{\rm SA} \cup \mathcal{V}_{\rm PA}$}{
            Observe the current state $o_{i,t}$ and $e_{j,i,t}$\;
        }

        \ForEach{agent $i \in \mathcal{V}_{\rm SA}$}{
            Generate $s_{i,t}$ according to Equation (13)\;
            Select action $a_{i,t}$ using policy $\pi_{\theta_{{\rm SA}_i}}(a_{i,t} | s_{i,t})$\;
        }

        \ForEach{agent $i \in \mathcal{V}_{\rm PA}$}{
            Generate $s_{i,t}$ according to Equation (13)\;
            Select action $a_{i,t}$ using policy $\pi_{\theta_{\rm PA}}(a_{i,t} | s_{i,t})$\;
        }

        Execute actions of multiple agents \( a_t = (a_{i,t})_{i \in \mathcal{V}_{\rm SA} \cup \mathcal{V}_{\rm PA}} \)\;

        \ForEach{agent $i \in \mathcal{V}_{\rm SA} \cup \mathcal{V}_{\rm PA}$}{
            Obtain reward $r_{i,t}$ and the next state $o_{i,t+1}$\;
            Store transition $(o_{i,t}, a_{i,t}, r_{i,t}, o_{i,t+1})$ in batch buffer.
        }
}
\end{algorithm}

\begin{algorithm}
\caption{Multi-Agent PPO Training}
\fontsize{9pt}{10pt}\selectfont
\ForEach{agent $i$}{
    Initialize policy network $\pi_{\theta_{{\rm SA}_i}}(a_{i,t} | s_{i,t})$, $\pi_{\theta_{\rm PA}}(a_{i,t} | s_{i,t})$ and value network $V_{\theta_{{\rm SA}_i}}(s_{i,t})$, $V_{\theta_{\rm PA}}(s_{i,t})$ parameters\;
    Initialize policy network optimizer and value network optimizer with parameters $\theta_{{\rm SA}_i}, \theta_{\rm PA}$\;
    Determine the edge feature among SAs $\left\{ e_{j,i,t} \mid i, j \in \mathcal{V}_{\rm SA} \right\}$\;
}
Set hyperparameters (e.g., learning rate, discount factor $\gamma$, total training steps, batch size, iteration scheme)\;

\For{cycle in total\_cycles}
{
    \For{iteration in training\_iterations}{
        Interact with the environment and collect data in the batch buffer using Algorithm \ref{algorithm1}\\
        \ForEach{agent $i \in \mathcal{V}_{\rm SA}$}{
                Generate $s_{i,t+1}$ according to Equation (13)\;
                Calculate advantage estimates $\hat{A}_{i,t}$ using Equation (20)\;
                Calculate PPO objective function $J_i(\theta_{{\rm SA}_i})$ using Equation (22)\;
                Update policy network parameters of $\theta_{{\rm SA}_i}$ by maximizing $J_i(\theta_{{\rm SA}_i})$ with the optimizer\;
                Update value network parameters of $\theta_{{\rm SA}_i}$ by minimizing the value function loss $L_i(\theta_{{\rm SA}_i})$, as defined in Equation (24)\;}}
    \For{iteration in training\_iterations}{
        Interact with the environment and collect data in the batch buffer using Algorithm \ref{algorithm1}\\
        \ForEach{agent $i \in \mathcal{V}_{\rm PA}$}{
                Generate $s_{i,t+1}$ according to Equation (13)\;
                Calculate advantage estimates $\hat{A}_{i,t}$ using Equation (21)\;
                Calculate PPO objective function $J_i(\theta_{\rm PA})$ using Equation (23)\;
                Update policy network parameters of $\theta_{\rm PA}$ by maximizing $J_i(\theta_{\rm PA})$ with the optimizer\;
                Update value network parameters of $\theta_{\rm PA}$ by minimizing the value function loss $L_i(\theta_{\rm PA})$, as defined in Equation (25)\;
            }
            }
}
\end{algorithm}

In the multi-agent PPO algorithm, the advantage estimate $\hat{A}_{i,t}$ is calculated as follows:
\begin{equation}
\hat{A}_{i,t} = r_{i,t} + \gamma \cdot V_{\theta_{{\rm SA}_i}}(s_{i,t+1}) - V_{\theta_{{\rm SA}_i}}(s_{i,t})\quad \forall i \in \mathcal{V}_{\rm SA} 
\end{equation}
\begin{equation}
\hat{A}_{i,t} = r_{i,t} + \gamma \cdot V_{\theta_{\rm PA}}(s_{i,t+1}) - V_{\theta_{\rm PA}}(s_{i,t})\quad \forall i \in \mathcal{V}_{\rm PA} 
\end{equation}
where $\hat{A}_{i,t}$ represents an estimator of the advantage function for agent $i$ at time step $t$; $V_{\theta_{{\rm SA}_i}}(s_{i,t}),V_{\theta_{\rm PA}}(s_{i,t})$ are the estimated value function for SA $i$ and PA, respectively; $\gamma$ is the discount factor.

The advantage function is used to guide policy updates, and the probability ratio $\rho_{i,t}(\theta_i)$ is employed to adjust the magnitude of policy updates. The clipping operation limits the step size of policy updates, ensuring the proximity of new and old policies and the stability of the learning process. The objective function for policy updates for agent $i$ is defined as:
\begin{equation}
\begin{aligned}
J_i(\theta_{{\rm SA}_i}) = \mathbb{E} \left[ \min (\rho_{{\rm SA}_i,t} \hat{A}_{i,t}, 
\text{clip}(\rho_{{\rm SA}_i,t}, 1-\epsilon, 1+\epsilon) \hat{A}_{i,t}) \right] \\ 
\quad \forall i \in \mathcal{V}_{\rm SA} 
\end{aligned}
\end{equation}

\begin{equation}
\begin{aligned}
J_i(\theta_{\rm PA}) = \mathbb{E} \left[ \min (\rho_{{\rm PA},t}\hat{A}_{i,t}, 
\text{clip}(\rho_{{\rm PA},t}, 1-\epsilon, 1+\epsilon) \hat{A}_{i,t}) \right] \\
\quad \forall i \in \mathcal{V}_{\rm PA} 
\end{aligned}
\end{equation}
where $J_i(\theta_{{\rm SA}_i}), J_i(\theta_{\rm PA})$ are the objective functions for policy updates of SA $i$ ($\forall i \in \mathcal{V}_{\rm SA} $) and PA $i$ ($\forall i \in \mathcal{V}_{\rm PA} $) respectively; $\rho_{{\rm SA}_i,t} , \rho_{{\rm PA},t} $ are the probability ratios of the current policy to the old policy for SA $i$ ($\forall i \in \mathcal{V}_{\rm SA} $) and PA $i$ ($\forall i \in \mathcal{V}_{\rm PA} $) respectively, that is, \(\rho_{{\rm SA}_i,t}= \frac{\pi_{\theta_{{\rm SA}_i}}(a_{i,t} \mid s_{i,t})}{\pi_{\theta'_{{\rm SA}_i}}(a_{i,t} \mid s_{i,t})}\),\(\rho_{{\rm PA},t} = \frac{\pi_{\theta_{\rm PA}}(a_{i,t} \mid s_{i,t})}{\pi_{\theta'_{\rm {PA}}}(a_{i,t} \mid s_{i,t})}\). $\pi_{\theta_{{\rm SA}_i}}(a_{i,t}|s_{i,t}), \pi_{\theta'_{{\rm SA}_i}}(a_{i,t}|s_{i,t})$ are the new and old policy functions of signal control agent $i$ ($\forall i \in \mathcal{V}_{\rm SA} $), indicating the probability of agent $i$ choosing action $a_{i,t}$ given state $s_{i,t}$ under the new and old policy parameters $\theta_{{\rm SA}_i}, \theta'_{{\rm SA}_i}$; $\pi_{\theta_{\rm PA}}(a_{i,t}|s_{i,t}), \pi_{\theta'_{{\rm PA}}}(a_{i,t}|s_{i,t})$ are the new and old policy functions of the PA; $\epsilon$ is a hyperparameter that controls the clipping range.

The value function loss for agent $i$ is defined as:
\begin{equation}
\begin{aligned}
L_i(\theta_{{\rm SA}_i}) = \mathbb{E} \left[ \left( r_{i,t} + \gamma V_{\theta_{{\rm SA}_i}}(s_{i,t+1}) - V_{\theta_{{\rm SA}_i}}(s_{i,t}) \right)^2 \right] \\
\quad \forall i \in \mathcal{V}_{\rm SA} 
\end{aligned}
\end{equation}

\begin{equation}
\begin{aligned}
L_i(\theta_{\rm PA}) = \mathbb{E} \left[ \left( r_{i,t} + \gamma V_{\theta_{\rm PA}}(s_{i,t+1}) - V_{\theta_{\rm PA}}(s_{i,t}) \right)^2 \right] \\
\quad \forall i \in \mathcal{V}_{\rm PA} 
\end{aligned}
\end{equation}
where $L_i(\theta_{{\rm SA}_i}), L_i(\theta_{\rm PA})$ are the value function losses for signal control agent $i$ ($\forall i \in \mathcal{V}_{\rm SA} $) and platoon management agent $i$ ($\forall i \in \mathcal{V}_{\rm PA} $) respectively.

\section{Simulation Experiments}
\subsection{Settings}
We use Simulation of Urban MObility (SUMO) \cite{krajzewicz2012recent}, an open-source package for microscopic traffic simulation, to build the environment. In this environment, SA and PA agents act as controllers, interacting through the TraCI (Traffic Control Interface) API to receive observations, obtain rewards, and take actions.

\begin{figure}[h!]
	\centering
		\includegraphics[scale=.3]{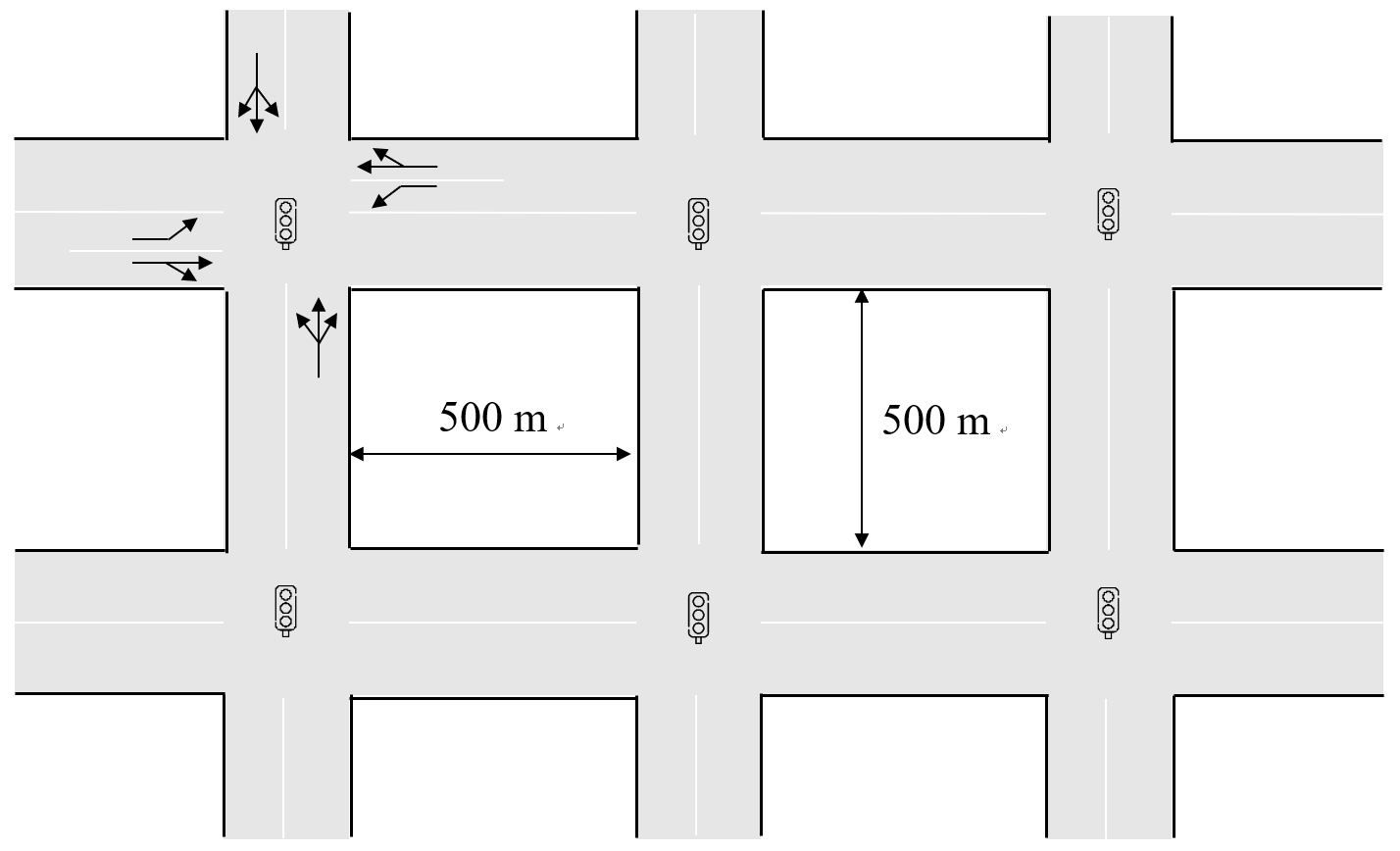}
	\caption{Road network for SUMO simulation with 6 OD pairs of platoons}
	\label{fig: network}
\end{figure}

The layout of the network and intersections is shown in Fig. \ref{fig: network}. The road network is a $2 \times 3$ grid with six signalized intersections, incorporating six SAs and a control pool of 100 PAs within the system.

The experiment setup involves various parameters including traffic demand, characteristics of vehicles and platoons such as speed limit and thresholds for platoon merging, as well as RL training parameters. These parameters are summarized in Table \ref{tab: parameters}.
\begin{table}[!ht]
\renewcommand\arraystretch{1.4}
\centering
\caption{Parameter Settings}
\begin{tabular}{ll}
\hline
\cline{1-2}
Description    & Values \\
\hline
Traffic arrival ratio (per intersection)                   & 1641 veh/h \\
Speed limit for solo vehicles                              & 35mph\\
Action space for PA $\mathcal{A}_{\rm PA}$ &[15, 20, 25,\\ (maximum allowable speed set) &  30, 35] mph\\
The maximum platoon size               &10 \\
The time headway threshold for platoon merge   &2s\\
The distance threshold for platoon merge   &2m\\
Desired minimum time headway for Krauss model & 1.5s\\
Desired minimum time headway for EIDM model & 1.0s\\
Speed deviation for vehicles in platoons &0.0\\
Speed deviation for solo vehicles &0.8\\
Control step &5s\\
Total cycles &12\\
Training iterations per cycle &100\\
Batch size &1000\\
\hline
\cline{1-2}
\label{tab: parameters}
\end{tabular}
\end{table}

To quantify the performance of the control policies, we use the following metrics:

\begin{itemize}
    \item \textbf{Average Travel Time:} The total travel time of all vehicles during the episode divided by the number of vehicles.
    \item \textbf{Average Fuel Consumption:} The total fuel consumption of all vehicles during the episode divided by the number of vehicles, measured using the HBEFA3 database \cite{keller2017hbefa}.
\end{itemize}

\subsection{Training Performance}
The training performance of the proposed method is illustrated in Fig. 6 and Fig. 7. The horizontal axis represents episodes, while the vertical axis shows average travel time and average fuel consumption respectively. Overall, performance improves significantly throughout the training process.

The proposed method alternately updates the policies of two types of agents. The background color indicates the type of agent being updated: blue for SA and green for PA. For each cycle, both SA and PA start from poor performance due to policy exploration. As the policies are updated, performance rapidly improves and gradually stabilizes.

As the learning process progresses, the worst and best values (always the starting and ending points) for each cycle generally decrease sequentially. Fig. 8 and Fig.9 present the average of the worst 20 values and the best 20 values for each cycle and the cumulative optimization percentage curve for each update. For instance, after the first round of SA updates, the average travel time reaches 244.4s and fuel consumption reaches 170.0g. In the second round, with the SA policy fixed and the PA policy updated, the travel time reaches 238.6s and fuel consumption reaches 163.5g. Compared to the first round results, travel time is optimized by 2.4\%, and fuel consumption is optimized by 3.8\%. The performance in the last cycle shows 6.4\% optimization in travel time and 8.5\% optimization in fuel consumption compared to the first cycle. This indicates that after updating the SA (PA) policy, the PA (SA) policy can be further updated to enhance traffic efficiency. The alternating training method enables the agents to learn each other's policies, thereby achieving coordination with better control performance.

\begin{figure}[ht!]
	\centering
		\includegraphics[scale=.25]{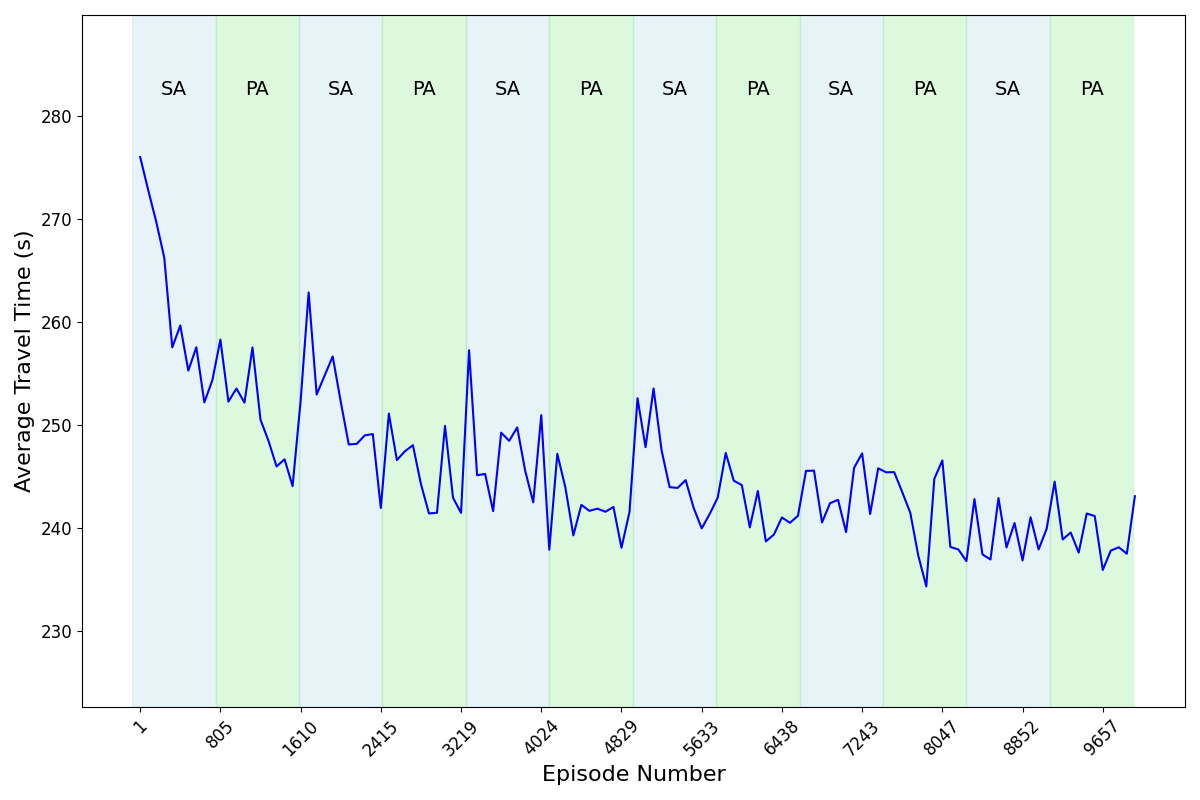}
	\caption{Average travel time throughout training}
	\label{fig: delay}
\end{figure}

\begin{figure}[ht!]
	\centering
		\includegraphics[scale=.25]{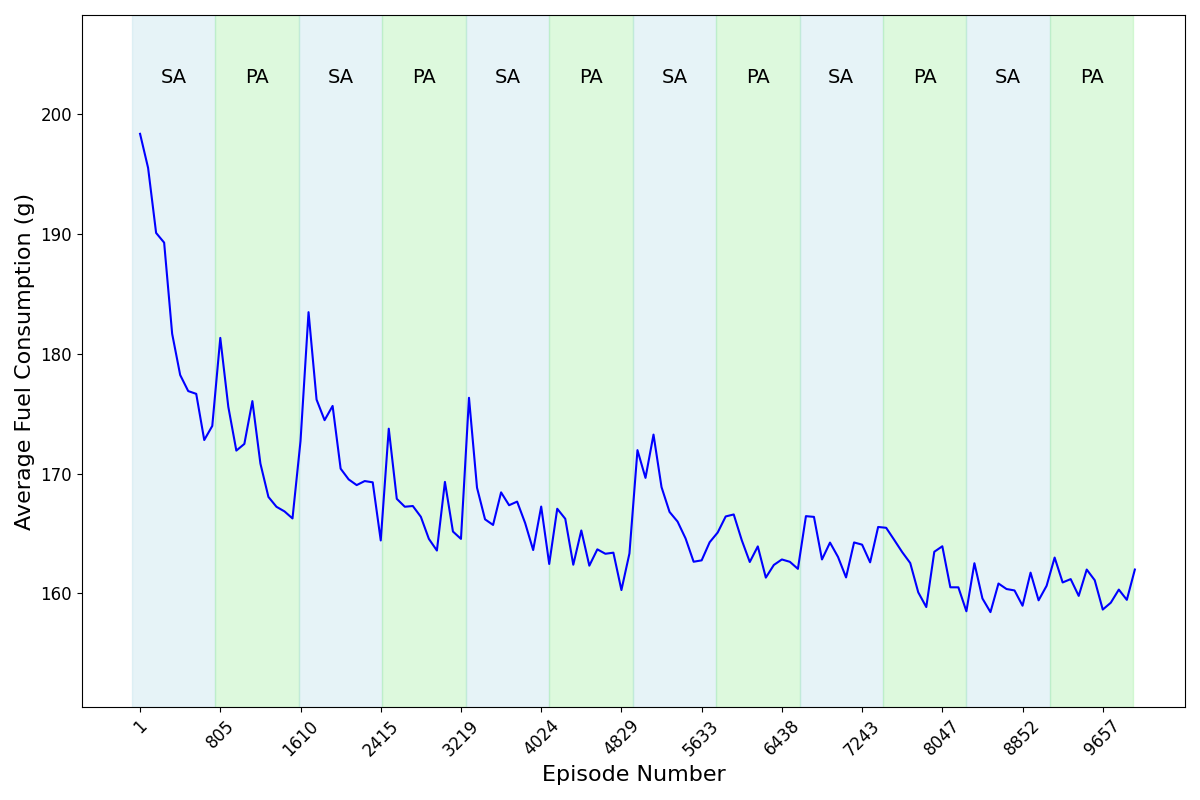}
	\caption{Average fuel consumption throughout training}
	\label{fig: fuelconsumption}
\end{figure}

\begin{figure}[ht!]
	\centering
		\includegraphics[scale=.5]{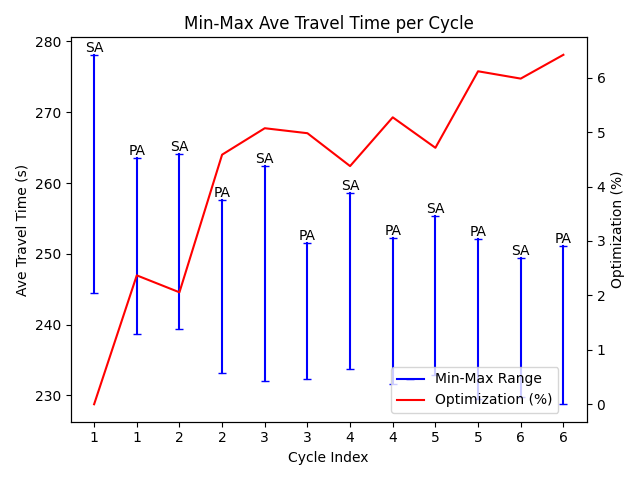}
	\caption{Min-Max Average Travel Time and Cumulative Optimization Percentage per Cycle}
	\label{fig: delay}
\end{figure}

\begin{figure}[ht!]
	\centering
		\includegraphics[scale=.5]{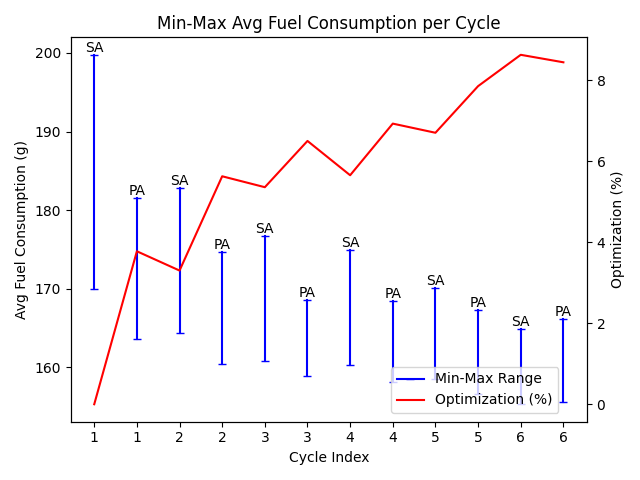}
	\caption{Min-Max Average fuel consumption and Cumulative Optimization Percentage per Cycle}
	\label{fig: fuelconsumption}
\end{figure}

\subsection{Performance Comparison}
To further investigate the performance of JointSP, we compare JointSP with the Multi-Agent Advantage Actor-Critic (MAA2C) method proposed by \cite{chu2019multi} and an adaptive signal control method called backpressure control\cite{zaidi2016back}. The control strategies discussed are as follows:

\begin{itemize}
    \item The proposed Joint SP method: This method promotes cooperation between SAs and PAs by sharing observations and some NN parameters within the GNN structure. Alternating optimization iteratively updates SA and PA policies, enabling agents to adapt to each other's strategy changes for better coordination and overall system performance.

    \item The MAA2C-based signal control method:  In this multi-agent system, each signal agent controls signal timings for an intersection, with some observations shared among adjacent intersections. Each SA's policy is trained using the A2C algorithm.

    \item Backpressure control: This method dynamically adjusts traffic signal timings based on real-time conditions. It calculates "pressure" at intersections by comparing vehicles in upstream and downstream lanes, then adjusts signals to reduce pressure and prioritize movement in congested directions to balance traffic flow and reduce congestion.
\end{itemize}

To evaluate the performance of the above control methods, we implement the trained policy into the environment for 4 episodes. The comparison results are shown in Table \ref{tab:performance_comparison}.
\begin{table}[h]
\renewcommand\arraystretch{1.4}
\centering
\caption{Evaluation of Trained Policies in Terms of Average Travel Time and Fuel Consumption}
\begin{tabular}{l p{2.5cm} p{2.5cm} }
\hline
\textbf{Metric} & \textbf{Average Travel\newline Time (s)} & \textbf{Average Fuel\newline Consumption (g)} \\
\hline
\textbf{JointSP} &  239.3 &  157.6 \\
\textbf{MA2C Signal Control} &  261.1 &  169.3 \\
\textbf{Backpressure Control} &  277.0 &  188.7 \\
\hline
\end{tabular}
\label{tab:performance_comparison}
\end{table}

Based on the data in Table \ref{tab:performance_comparison}, the JointSP method, which uses cooperative control of SA and PA, outperforms the adaptive signal control methods. JointSP more effectively reduces both average travel time (by 8.4\% to 13.6\%) and fuel consumption (by 6.9\% to 16.5\%) compared to MA2C Signal Control and Backpressure Control. By iteratively learning and adapting strategies between signal agents (SAs) and platoon agents (PAs), JointSP achieves a synergistic approach, significantly improving system performance.
 
\subsection{Sensitivity Analysis for Traffic Demand and CAV rate}
To evaluate the robustness of JointSP, we conducted a sensitivity analysis involving varying traffic demands. We implemented the trained policies in road networks with different traffic demands, and the results are shown in Fig. 10. As traffic demand increases, both fuel consumption and average travel time rise significantly. When the traffic demand is 0.8 times the training traffic demand, the upward trend slows down. At 1.2 to 1.4 times the traffic demand, fuel consumption and travel time are similar to those at 1.0 times the traffic demand, indicating effective control. This demonstrates that JointSP's coordination can effectively adapt to different levels of traffic demand, showcasing its robustness and adaptability in handling varying traffic conditions.

In our proposed mechanism, each solo vehicle in the environment can potentially become part of a platoon, assuming all vehicles are CAVs. To evaluate the effectiveness of JointSP under different CAV penetration rates, we incorporated human-driven vehicles, which have characteristics similar to solo vehicles but cannot form platoons. This allowed us to implement and evaluate the trained joint policies in environments with varying CAV penetration rates.

As shown in Fig. 11, the results indicate that our model is most effective when the CAV penetration rate is particularly high. This is because platoons need a high CAV penetration rate to exist consistently and extensively. If there is a significant proportion of HVs mixed in with the CAVs, it becomes difficult to form effective PAs. When the CAV rate is below 0.4, it is even challenging to form platoons of more than three vehicles in the network, making it difficult to achieve optimal control performance. In our future work, we aim to develop a solution that can effectively guide platoons through intersections even with low CAV penetration rates, utilizing a small number of CAVs to lead the platoons.

\begin{figure}[ht!]
    \centering
    \includegraphics[scale=.4]{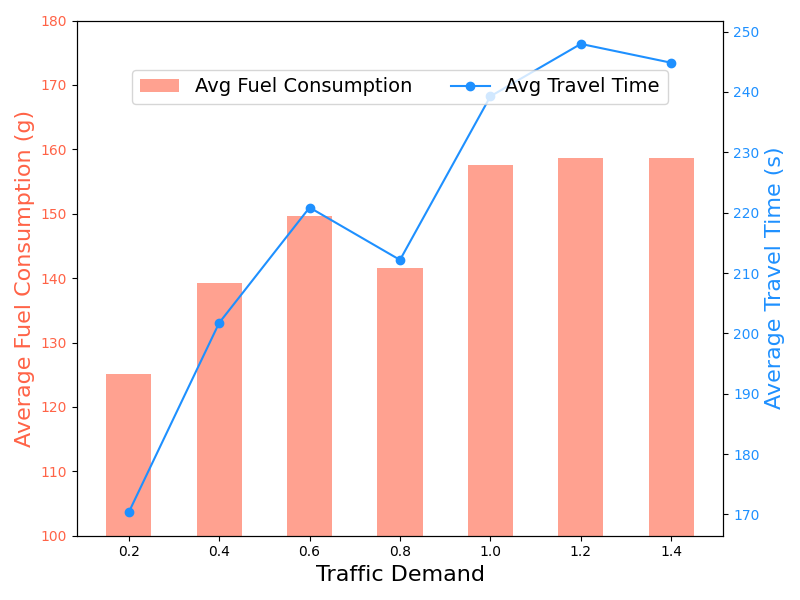}
    \caption{System performance under varying traffic demands}
    \label{fig:sensitivity_analysis_diff_demand}
\end{figure}

\begin{figure}[ht!]
    \centering
    \includegraphics[scale=.4]{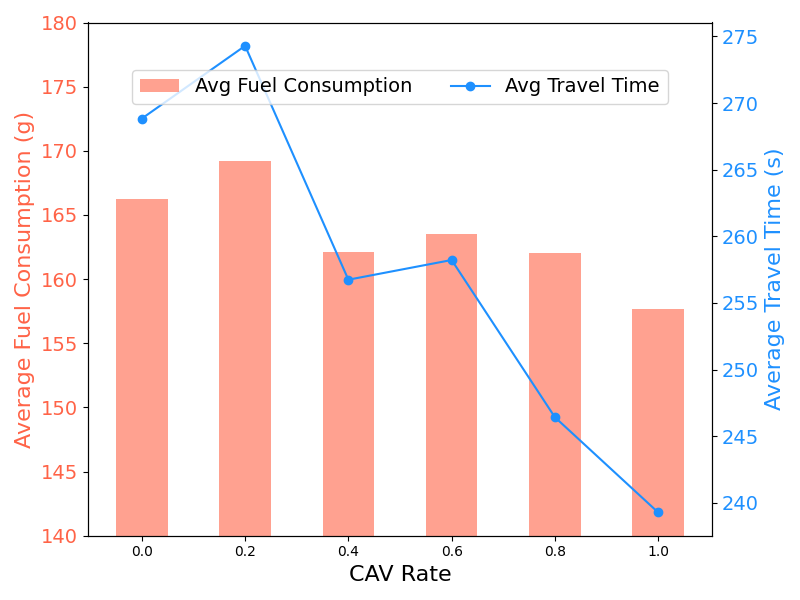}
    \caption{System performance under varying CAV rate}
    \label{fig:sensitivity_analysis_diff_CAV_rate}
\end{figure}

\section{Conclusion}
This paper presents JointSP, a cooperative graph-based multi-agent reinforcement learning framework to coordinate platoon and signal control strategies simultaneously in a possible way. It allows platoons to navigate signalized intersections with minimal or no stops. Both SAs and PAs are carefully designed to accommodate the inherent physical and behavioral disparities among these agents. Additionally, they follow distinct training processes based on their geographical features and characteristics, with SAs utilizing the DTDE paradigm and PAs employing the CTDE paradigm. To facilitate coordination and communication between heterogeneous agents, the HGMARL framework is integrated with GNN to enhance their ability to make optimal decisions collectively.

Alternating optimization improves SA (PA)'s own policy and adapts to PA's (SA's) policy, thereby achieving coordination and enhancing performance. Compared to MAA2C-based signal control and backpressure control, JointSP demonstrates better performance in terms of travel time and fuel consumption.  

However, it is important to acknowledge the limitations of our study. While we have demonstrated the effectiveness of our proposed method, a comprehensive comparison with state-of-the-art RL techniques requires further exploration. Also, the applicability to lower CAV penetration rate and much larger city-wide road networks needs to be further studied.
%\bibliography{reference}

\begin{thebibliography}{99}
\bibitem{liang2019deep}
X. Liang, X. Du, G. Wang, and Z. Han, ``A deep reinforcement learning network for traffic light cycle control,'' \textit{IEEE Transactions on Vehicular Technology}, vol. 68, no. 2, pp. 1243--1253, 2019.

\bibitem{lee2019reinforcement}
J. Lee, J. Chung, and K. Sohn, ``Reinforcement learning for joint control of traffic signals in a transportation network,'' \textit{IEEE Transactions on Vehicular Technology}, vol. 69, no. 2, pp. 1375--1387, 2019.

\bibitem{abdoos2021hierarchical}
M. Abdoos and A. L. C. Bazzan, ``Hierarchical traffic signal optimization using reinforcement learning and traffic prediction with long-short term memory,'' \textit{Expert Systems with Applications}, vol. 171, p. 114580, 2021.

\bibitem{zhang2021cooperative}
Y. Zhang, Y. Zhou, H. Lu, and H. Fujita, ``Cooperative multi-agent actor-critic control of traffic network flow based on edge computing,'' \textit{Future Generation Computer Systems}, vol. 123, pp. 128--141, 2021.

\bibitem{yang2021ihg}
S. Yang, B. Yang, Z. Kang, and L. Deng, ``IHG-MA: Inductive heterogeneous graph multi-agent reinforcement learning for multi-intersection traffic signal control,'' \textit{Neural Networks}, vol. 139, pp. 265--277, 2021.

\bibitem{INRIX}
B. Pishue, ``2022 INRIX Global Traffic Scorecard,'' INRIX Research, 2023.

\bibitem{Fast_Facts}
U.S. Environmental Protection Agency, ``Fast facts: U.S. transportation sector greenhouse gas emissions 1990-2017,'' \url{https://www.epa.gov/greenvehicles/fast-facts-transportation-greenhouse-gas-emissions}, 2020.

\bibitem{krajzewicz2012recent}
D. Krajzewicz, J. Erdmann, M. Behrisch, and L. Bieker, ``Recent development and applications of SUMO-Simulation of Urban MObility,'' \textit{International Journal on Advances in Systems and Measurements}, vol. 5, no. 3\&4, 2012.

\bibitem{desjardins2011cooperative}
C. Desjardins and B. Chaib-Draa, ``Cooperative adaptive cruise control: A reinforcement learning approach,'' \textit{IEEE Transactions on Intelligent Transportation Systems}, vol. 12, no. 4, pp. 1248--1260, 2011.

\bibitem{buechel2018deep}
M. Buechel and A. Knoll, ``Deep reinforcement learning for predictive longitudinal control of automated vehicles,'' in \textit{Proc. 21st International Conference on Intelligent Transportation Systems (ITSC)}, pp. 2391--2397, 2018.

\bibitem{chu2019model}
T. Chu and U. Kalabi{\'c}, ``Model-based deep reinforcement learning for CACC in mixed-autonomy vehicle platoon,'' in \textit{Proc. IEEE 58th Conference on Decision and Control (CDC)}, pp. 4079--4084, 2019.

\bibitem{lu2021sharing}
S. Lu, Y. Cai, L. Chen, H. Wang, X. Sun, and Y. Jia, ``A sharing deep reinforcement learning method for efficient vehicle platooning control,'' \textit{IET Intelligent Transport Systems}, 2021.

\bibitem{chen2020platooning}
C. Chen, Y. Zhang, M. R. Khosravi, Q. Pei, and S. Wan, ``An intelligent platooning algorithm for sustainable transportation systems in smart cities,'' \textit{IEEE Sensors Journal}, 2020.

\bibitem{li2017training}
Z. Li, T. Chu, I. V. Kolmanovsky, and X. Yin, ``Training drift counteraction optimal control policies using reinforcement learning: An adaptive cruise control example,'' \textit{IEEE Transactions on Intelligent Transportation Systems}, vol. 19, no. 9, pp. 2903--2912, 2017.

\bibitem{zhou2019development}
M. Zhou, Y. Yu, and X. Qu, ``Development of an efficient driving strategy for connected and automated vehicles at signalized intersections: A reinforcement learning approach,'' \textit{IEEE Transactions on Intelligent Transportation Systems}, vol. 21, no. 1, pp. 433--443, 2019.


\bibitem{lei2022deep}
L. Lei, T. Liu, K. Zheng, and L. Hanzo, ``Deep reinforcement learning aided platoon control relying on V2X information,'' \textit{IEEE Transactions on Vehicular Technology}, vol. 71, no. 6, pp. 5811--5826, 2022.

\bibitem{shi2023deep}
H. Shi, D. Chen, N. Zheng, X. Wang, Y. Zhou, and B. Ran, ``A deep reinforcement learning based distributed control strategy for connected automated vehicles in mixed traffic platoon,'' \textit{Transportation Research Part C: Emerging Technologies}, vol. 148, p. 104019, 2023.

\bibitem{berbar2022reinforcement}
A. Berbar, A. Gastli, N. Meskin, M. A. Al-Hitmi, J. Ghommam, M. Mesbah, and F. Mnif, ``Reinforcement learning-based control of signalized intersections having platoons,'' \textit{IEEE Access}, vol. 10, pp. 17683--17696, 2022.

\bibitem{bettini2023heterogeneous}
M. Bettini, A. Shankar, and A. Prorok, ``Heterogeneous multi-robot reinforcement learning,'' \textit{arXiv preprint arXiv:2301.07137}, 2023.

\bibitem{christianos2021scaling}
F. Christianos, G. Papoudakis, M. A. Rahman, and S. V. Albrecht, ``Scaling multi-agent reinforcement learning with selective parameter sharing,'' in \textit{Proc. International Conference on Machine Learning}, pp. 1989--1998, 2021.

\bibitem{lowe2017multi}
R. Lowe, Y. I. Wu, A. Tamar, J. Harb, P. Abbeel, and I. Mordatch, ``Multi-agent actor-critic for mixed cooperative-competitive environments,'' \textit{Advances in Neural Information Processing Systems}, vol. 30, 2017.

\bibitem{wang2020roma}
T. Wang, H. Dong, V. Lesser, and C. Zhang, ``ROMA: Multi-agent reinforcement learning with emergent roles,'' \textit{arXiv preprint arXiv:2003.08039}, 2020.

\bibitem{yen2022deep}
C.-C. Yen, H. Gao, and M. Zhang, ``Deep reinforcement learning based platooning control for travel delay and fuel optimization,'' in \textit{Proc. IEEE 25th International Conference on Intelligent Transportation Systems (ITSC)}, pp. 737--742, 2022.

\bibitem{schulman2017proximal}
J. Schulman, F. Wolski, P. Dhariwal, A. Radford, and O. Klimov, ``Proximal policy optimization algorithms,'' \textit{arXiv preprint arXiv:1707.06347}, 2017.

\bibitem{chen2020toward}
C. Chen, H. Wei, N. Xu, G. Zheng, M. Yang, Y. Xiong, K. Xu, and Z. Li, ``Toward a thousand lights: Decentralized deep reinforcement learning for large-scale traffic signal control,'' in \textit{Proc. AAAI Conference on Artificial Intelligence}, vol. 34, no. 4, pp. 3414--3421, 2020.

\bibitem{xu2021hierarchically}
B. Xu, Y. Wang, Z. Wang, H. Jia, and Z. Lu, ``Hierarchically and cooperatively learning traffic signal control,'' in \textit{Proc. AAAI Conference on Artificial Intelligence}, vol. 35, no. 1, pp. 669--677, 2021.

\bibitem{keller2017hbefa}
M. Keller, S. Hausberger, C. Matzer, P. W{\"u}thrich, and B. Notter, ``HBEFA Version 3.3,'' \textit{Background documentation, Berne}, vol. 12, 2017.

\bibitem{sun2023hierarchical}
Q. Sun, L. Zhang, H. Yu, W. Zhang, Y. Mei, and H. Xiong, ``Hierarchical reinforcement learning for dynamic autonomous vehicle navigation at intelligent intersections,'' in \textit{Proc. 29th ACM SIGKDD Conference on Knowledge Discovery and Data Mining}, pp. 4852--4861, 2023.

\bibitem{chu2019multi}
T. Chu, J. Wang, L. Codec{\`a}, and Z. Li, ``Multi-agent deep reinforcement learning for large-scale traffic signal control,'' \textit{IEEE Transactions on Intelligent Transportation Systems}, vol. 21, no. 3, pp. 1086--1095, 2019.

\bibitem{feng2024hierarchical}
P. Feng, J. Liang, S. Wang, X. Yu, R. Shi, and W. Wu, ``Hierarchical consensus-based multi-agent reinforcement learning for multi-robot cooperation tasks,'' \textit{arXiv preprint arXiv:2407.08164}, 2024.

\bibitem{zaidi2016back}
A. A. Zaidi, B. Kulcs{\'a}r, and H. Wymeersch, ``Back-pressure traffic signal control with fixed and adaptive routing for urban vehicular networks,'' \textit{IEEE Transactions on Intelligent Transportation Systems}, vol. 17, no. 8, pp. 2134--2143, 2016.


\end{thebibliography}

\end{document}